# BETA-RANK: A ROBUST CONVOLUTIONAL FILTER PRUNING METHOD FOR IMBALANCED MEDICAL IMAGE ANALYSIS

**Morteza Homayounfar,**
Department of Diagnostic Radiology,
Li Ka Shing Faculty of Medicine,
The University of Hong Kong,
mohofar@hku.hk

**Mohamad Koohi-Moghadam,**
Division of Applied Oral Sciences
& Community Dental Care,
Faculty of Dentistry,
The University of Hong Kong,
koohi@hku.hk

**Reza Rawassizadeh,**
Department of Computer Science,
Metropolitan College,
Boston University,
rezar@bu.edu

**Varut Vardhanabhuti[1],**
Department of Diagnostic Radiology,
Li Ka Shing Faculty of Medicine,
The University of Hong Kong,
varv@hku.hk

## ABSTRACT

As deep neural networks include a high number of parameters and operations, it can be a challenge to implement these models on devices with limited computational resources. Despite the development of novel pruning methods toward resource-efficient models, it has become evident that these models are not capable of handling "imbalanced" and "limited number of data points". We proposed a novel filter pruning method by considering the input and output of filters along with the values of the filters that deal with imbalanced datasets better than others. Our pruning method considers the fact that all information about the importance of a filter may not be reflected in the value of the filter. Instead, it is reflected in the changes made to the data after the filter is applied to it. In this work, three methods are compared with the same training conditions except for the ranking values of each method, and 14 methods are compared from other papers. We demonstrated that our model performed significantly better than other methods for imbalanced medical datasets. For example, when we removed up to 58% of FLOPs for the IDRID dataset and up to 45% for the ISIC dataset, our model was able to yield an equivalent (or even superior) result to the baseline model. To evaluate FLOP and parameter reduction using our model in real-world settings, we built a smartphone app, where we demonstrated a reduction of up to 79% in memory usage and 72% in prediction time. All codes and parameters for training different models are available at https://github.com/mohofar/Beta-Rank

## INTRODUCTION

The advancement of convolutional neural networks (CNNs) has led to significant breakthroughs in computer vision tasks [1], [2], [3] and they have been widely applied in various fields [4], [5], [6], including medical image analysis. Training a deep learning model with a customized architecture tailored to a specific task requires sufficient knowledge and experience in network design, which is challenging. Consequently, most researchers prefer to use pre-trained models and fine-tune them through transfer learning. Occasionally, the capacity of neural networks may not always align with newly developed tasks, leading to additional parameters and, consequently, additional computation costs after training. In situations with limited network availability or due to the privacy of data, the model is recommended to deploy the model on a device [7]. Such models face various resource constraints, including computation power, memory, and battery life [8]. Besides, in some real-time medical applications where CNNs are deployed in mobile environments (e.g. bedside), models must be implemented on edge devices while maintaining optimal performance [9], [10] with computations occurring in real-time or close to real-time operating within the constraint of these devices.

---

[1] Corresponding author

While with the advancement of deep learning models, the number of parameters and floating-point operations (FLOPs) is growing, and it is becoming difficult to implement these models on devices or in real-time applications. For example, ResNet50 [2], with 4.1 billion FLOPs [11] on average, and GPT-3 [12] with approximately 175 billion parameters, are two examples of common models that demand substantial computational resources to function. In this context, developing more resource-efficient models is crucial, necessitating the adoption of pruning techniques and minimizing the computational FLOPs of the model without compromising accuracy.

Two common approaches for pruning convolutional networks are weight pruning [13], [14] and filter pruning [15], [16], [17], [18], [19]. Weight pruning techniques involve adjusting the weight values of filters by assigning zero to less significant parameters, whereas filter pruning methods attempt to delete the entire filter. The filter pruning method is preferable to weight pruning as it enables the removal of entire low-ranked filters and optimization of the whole model while weight pruning is limited to specific software or hardware in real-world applications [20], [21].

There are many papers regarding filter pruning using different methods. $L_1$-Norm [22] is a widely-used model for filter pruning that indicates filter importance by the magnitude of filter values. In the Thinet method by Luo et al. [15], the authors suggest that the statistical information of layer i+1 can be used to prune filters in the i-th layer from an optimization perspective. Although effective, the computational cost of filter ranking may be high if an estimation algorithm is employed for each layer. The Soft Filter Pruning method [16] improves upon this by using an L2-Norm method and an iterative process for pruning and training filters, but potential bias in the training may compromise the approach's effectiveness. Meng et al.'s Stripe-Wise Pruning (SWP) [18] leverages weight and filter pruning with an additive filter called Filter Skeleton (FS), but this customization may not be easily applicable to different architectures. He et al. [19] explore the geometric median instead of norm-based ranking methods, focusing on filters with smaller values that play a critical role in feature representation. Hrank [11] uses Singular Value Decomposition (SVD) to rank filters, outperforming previous methods in various datasets. FilterSketch [23] efficiently reduces the complexity of pre-trained deep neural networks while preserving their information by formulating the pruning problem as a matrix sketch problem. Redundant Feature Pruning (RFP) [24] presents an efficient technique for pruning deep and wide convolutional neural network models by eliminating redundant features based on their differentiation and relative cosine distances in the feature space. Yu et al. [25] introduce the Neuron Importance Score Propagation (NISP) algorithm, which prunes CNNs by jointly considering all layers to minimize the reconstruction error in the final response layer. Lin et al. [26] propose an effective structured pruning approach for CNNs that jointly prunes filters and other structures in an end-to-end manner, overcoming the limitations of existing multi-stage, layer-wise methods. ABCPruner [27] is a new channel pruning method for deep neural networks based on the artificial bee colony (ABC) algorithm. It efficiently finds the optimal pruned structure by limiting the preserved channels to a specific space and using the ABC algorithm to solve the optimization problem. Global channel pruning (GCP) [28] introduces Performance-Aware Global Channel Pruning (PAGCP), a framework for multitask model compression that addresses task mismatch and filter interaction issues in multitask pruning. Lin et al. [29] present CLR-RNF, a novel filter-level network pruning method that identifies a "long-tail" pruning problem in magnitude-based weight pruning methods and proposes a computation-aware measurement for individual weight importance. CHIP [30] proposes an efficient filter pruning method using channel independence, which measures correlations among different feature maps. Sparse Structure Selection (SSS) [31] introduces a simple and effective framework to learn and prune deep CNNs in an end-to-end manner, using scaling factors and sparsity regularizations. Zhao et al. [32] propose a variational Bayesian scheme for pruning convolutional neural networks (CNNs) at the channel level, which improves computation efficiency by eliminating the need for re-training and can be easily implemented as a standalone module in existing deep learning packages.

Lin et al. [33] and Blakeney et al. [34] address the issue of bias in deep neural networks during pruning, with Lin et al. proposing the FairGRAPE pruning method to minimize the impact on different sub-groups and hidden biases, while Blakeney et al. focus on preventing algorithmic bias in pruned networks. However, these studies have not been evaluated on widely used datasets like Cifar10 to compare against popular models, and they require larger datasets for optimal performance.

Besides, there are some limitations in the previous works. In some cases [11], [15], [22], the most promising results were presented in conjunction with a newly developed training procedure and hyperparameters, making it difficult to determine whether the improvement in results was due to filter pruning or the use of a particular hyperparameter. In addition, the results of earlier methods were only reported on benchmark datasets [11], [15], [22], [34] while it is crucial to evaluate the methods on challenging datasets, such as medical images, which often have "fewer training samples" and are frequently "imbalanced". A pruning strategy may perform efficiently on well-balanced standard datasets such as CIFAR10, but it has limitations when applied to challenging datasets. To the best of our knowledge, no prior research has undertaken a complete analysis of both benchmark and imbalanced real-world datasets with few labels, such as medical images. We address the problems in our work and present a new method.

Here, we implemented a redesigned pruning strategy that can handle datasets with imbalanced data and small training samples. Our model outperformed other methods by a significant margin on medical image datasets, which have small and imbalanced training samples. It also performed well on a benchmark database, demonstrating its generalizability. We demonstrated on GradCam analysis that our model was able to localize to specific and relevant regions similar to baseline unpruned techniques whilst also maintaining good performance. Finally, pruned models demonstrate promising improvements in time and memory usage during execution on an Android phone.

## METHOD

### A. Notation

Given two layers of a CNN model, namely $l_i$ and $l_{i+1}$, each layer is fed with $N$ samples and each sample is specified by $j$. The size of layers is represented by $(n, m, h)$, serving as an example of a 2D model. A region $x$ specified on each image using dash lines, which shows the region where the convolution operation is applied based on a predefined filter size. The size of $x$ for the input layer ($x_i^j$) equals the filter size ($f_{i_k}$) and the third dimension of filters equals the third dimension of layer $l_i$ ($h_i = h'_i$). However, all dimensions of $x$ for the output layer ($l_{i+1}^j$) must be one as a result of convolution operation. An arbitrary number of filters is used and ascertained using variable $k$. Finally, the rank of filter number k for layer i is defined as $R_{i_k}$ (Figure 1).

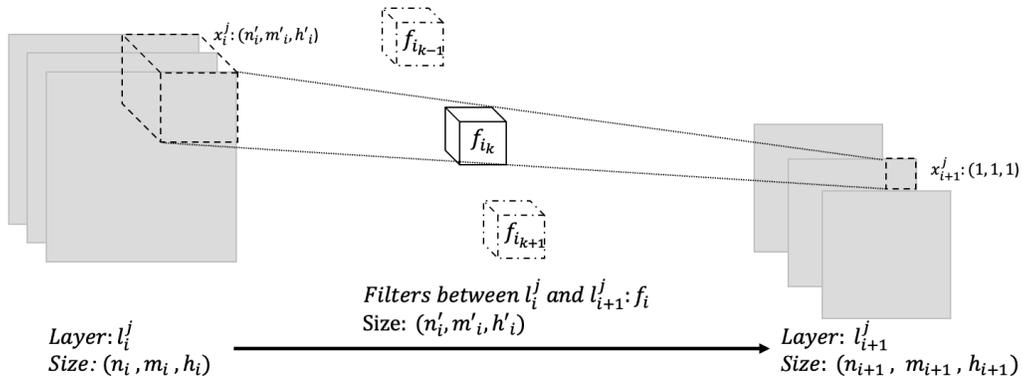

*Fig. 1. Two sample layers of a convolutional network for pruning.*

## B. β Rank

L1-Norm [22] is one of the popular and simple methods for filter ranking. The following equation describes it:

$$R_{i_k}^{L1} = \sum |f_{i_k}| \quad (1)$$

Equation 1 used larger values in filters to show the importance. Thus, after the calculation of $R_{i_k}^{L1}$ for each filter, we can sort and rank them. However, this is a partially true assumption on some occasions. Because in addition to the sum of data which is utilized by the L1-Norm method, the standard deviation of the input and output of filters is important, and when we have two variables to compare the filter, one variable might make a big difference that we are not aware of it. This issue is more likely to occur when we have a limited dataset or a dataset that has a bias in training data which is an indispensable part of datasets in the real world. As a resolution to mitigate this issue, we compact the information into only one equation by fixing another one, and this is the cornerstone of our method.

Assume the standard deviation of the input layer as $\sigma_i^p$ and the output layer as $\sigma_{i+1}^p$. The standard deviation is calculated for the position of $p$ which corresponds to the dash-line parts in Figure 1. For example, Equation 2 presents the standard deviation of layer $i + 1$.

$$\sigma_{i+1}^p = \sqrt{\frac{\sum (x_{i+1} - \mu_{i+1})^2}{N}}; \quad (2)$$
$$p = \{(a,b) \in \mathbb{z} \,|\, 1 \leq a \leq N_1, 1 \leq b \leq N_2\}$$

$\mu_{i+1}$ is the mean of the region ($x_{i+1}$) and $N$ is the number of samples in a batch that we feed to the model. The number of possible $\sigma_i$ that can be calculated depending on the size of the data and stride ($s$), as presented in Equation 3:

$$(N_1, N_2) = \left[\frac{(n_i, m_i) - (n_i', m_i')}{s}\right] + 1 \quad (3)$$

Finally, after calculating $N_1 \times N_2$ of $\sigma_{i+1}^p$ and averaging all of them, a final value can be achieved for this layer over a different number of samples. The same calculation can be accomplished for the other layers as we have an input and output pair for each layer.

For each filter, there is a pair of ($\sigma_i, \sigma_{i+1}$) based on the input of the filter and generated output after applying the filter. We can calculate the values of filters using Equation 4 which considers standard deviation as well:

$$f_{i_k}' \triangleq f_{i_k} * \frac{\sigma_{i+1}}{\sigma_i} \quad (4)$$

By multiplying the standard deviation fraction, we can make sure the filtering operations have implicitly all information in its amplitude, and comparing ranking the filters in L1-Norm, this method has more information on the fed data. Therefore, the overall equation of ranking using our model called $\beta$ is presented as Equation 5:

$$R_{i_k}^{\beta} = \sum |f_{i_k}'| = R_{i_k}^{L1} * \left|\frac{\sigma_{i+1}^p}{\sigma_i^p}\right| = R_{i_k}^{L1} * \sqrt{\frac{\sum_{j=1}^{N}(x_{i+1}^j - \mu_{i+1})^2}{\sum_{j=1}^{N}(x_i^j - \mu_i)^2}} = R_{i_k}^{L1} * \beta \quad (5)$$

The following pseudocode shows the procedure of our structural pruning step by step:

---
**Algorithm 1:** Procedure of Beta-Rank and fine-tuning
---
**Input**: A random batch of data for ranking filters ($I_{prn}$)
        Data for fine-tuning ($I_{org}$)
**Given**: $PR = [pr_1, pr_2, ..., pr_N]$; $0 < pr_n < 1$, $N = layer\ number$
**Initialize:** Original pretrained model to be pruned ($M_{org}$)
**For** $pr_n$ in PR:
        Calculate $R_{i_k}^{\beta}$ for each filter based on $M_{org}(I_{prn})$
Select top $(1 - PR)\%$ filters from Sorted $R_{i_k}^{\beta}$
Construct $M_{prn}$, based on $M_{org}$ and selected filters
**Initialize** $M_{prn}$ based on the weights of $M_{org}$
**For** epochs:
        Update $M_{prn}(I_{org})$
**Output:** $M_{prn}$

---

As described in the pseudocode, we construct a new model with fewer filters ($M_{prn} \subset M_{org}$) with the same architecture as the baseline model. The structural pruning leads to a low-parameters and low-FLOPs model that improves the speed of the prediction in the real-world setting.

## C. β Analysis

Equation 5 is composed of two parts of L_1-Norm value and $\beta$ fraction. The L1-Norm part assigns higher importance to more significant features [22] irrespective of the variability in the data and does not account for the impact of individual samples on ranking. In contrast, the $\beta$ fraction uses samples to consider their effect on datasets that exhibit imbalances. In the following, we explore the effect of $\beta$ fraction in the context of bias trained models.

In the context of biased model caused by samples during the training of a model, we have major and minor classes. The minor classes are underrepresented and can be considered unnecessary variations just like noise as they cannot change the loss function of a model very much in contrast with major classes. Thus, those filters that capture features of minor classes can be ranked lower by the L1-Norm method, but they might have valuable information about underrepresented classes.

The $\beta$ fraction aims to quantify a filter's capacity to discern small variations within input data. A larger $\beta$ fraction signifies that a filter amplifies the disparity between input and output layers, thus indicating that the filter exhibits a heightened sensitivity to nuanced differences in the input data. This heightened sensitivity can potentially facilitate the effective capture of underrepresented features. In scenarios involving minority classes or underrepresented features, specific filters may exhibit superior efficacy in discerning rare features compared to others. By employing the $\beta$ fraction as a metric to gauge filter importance, we prioritize filters that exhibit an increased sensitivity to minute differences within the input data. Consequently, we can improve the model's ability to recognize and represent underrepresented features effectively. Upon pruning filters by considering their $\beta$ fractions, we retain filters exhibiting higher $\beta$ fractions while discarding those with lower $\beta$ fractions. As a result, the pruned network becomes more adept at capturing underrepresented features, as the remaining filters are more sensitive to subtle variations in the input data. This approach can potentially lead to enhanced model performance in tasks involving minority classes or rare features, as the network is better equipped to handle these complexities. In the following, we present the reasons more technically.

Given the $R_{i_k}^{\beta}$ as the product of $R_{i_k}^{L1}$ and $\beta$, consider two sets of filters in layer $l$ of a network, $F_{major}$ and $F_{minor}$, which capture features from major (overrepresented) and minor (underrepresented) classes, respectively:

$$F_{major} = \{f \in F_l \mid f \text{ captures features of major classes}\} \quad (6)$$
$$F_{minor} = \{f \in F_l \mid f \text{ captures features of minor classes}\} \quad (7)$$

Now, let's define the average L1-Norm and $\beta$ fraction for filters in $F_{major}$ and $F_{minor}$ to compare them for different scenarios:

$$L1_{major} = \left(\frac{1}{|F_{major}|}\right) * \sum_{f \in F_{major}} L1(f) \quad (8)$$

$$L1_{minor} = \left(\frac{1}{|F_{minor}|}\right) * \sum_{f \in F_{major}} L1(f) \quad (9)$$

$$\beta_{major} = \left(\frac{1}{|F_{major}|}\right) * \sum_{f \in F_{major}} \left|\frac{\sigma_f(x_{i+1})}{\sigma_f(x_i)}\right| \quad (10)$$

$$\beta_{minor} = \left(\frac{1}{|F_{minor}|}\right) * \sum_{f \in F_{minor}} \left|\frac{\sigma_f(x_{i+1})}{\sigma_f(x_i)}\right| \quad (11)$$

Typically, in imbalanced trained models, we expect that: $L1_{major} > L1_{minor}$ and $\beta_{major} < \beta_{minor}$. This is because filters capturing major classes are likely to have larger weight magnitudes, while filters capturing minor classes are expected to increase the standard deviation of input data in their output as they might be more sensitive to variations in the input data. Considering the $R_{i_k}^{\beta}$ equation, we can rewrite these as:

$$BetaRank_{major} = \left(\frac{1}{|F_{major}|}\right) * \sum_{f \in F_{major}} \left(L1(f) * \left|\frac{\sigma_f(x_{i+1})}{\sigma_f(x_i)}\right|\right) \quad (12)$$

$$BetaRank_{minor} = \left(\frac{1}{|F_{minor}|}\right) * \sum_{f \in F_{minor}} \left(L1(f) * \left|\frac{\sigma_f(x_{i+1})}{\sigma_f(x_i)}\right|\right) \quad (13)$$

Equations 12 and 13 show, we can have a balanced ranking if we combine two methods. For major classes, L1-Norm has larger values whereas beta fraction has smaller values and vice versa for minor classes. Overall, the Beta-Rank method aims to prioritize filters that capture both major and minor classes' important features, leading to a more balanced and stable model for handling imbalanced datasets.

## RESULTS

### D. *Datasets*

We evaluated our method on four different datasets. The first evaluation was performed on CIFAR-10 [35] and CIFAR-100 [35] with 50,000 training and 10,000 validation samples for each dataset with balanced-class labels. For further evaluation of challenging medical datasets which are useful for on-device usage, two datasets including the skin lesion dataset from the ISIC2017 challenge [36] and the diabetic macular edema severity grade from the IDRiD challenge [37] were selected. We selected fundus and skin cancer images because there were prior research works [38], [39] that evaluated the use of smartphones in underrepresented communities for skin and eye diseases. In real-world datasets, the two most common challenges are of "limited number of samples" and "imbalanced distribution" [40] and we addressed them in the pruning process using these datasets particularly because they demonstrate these characteristics. We summarize the information about our experimental datasets in Table 1.

Table 1: Datasets Description And Class Distribution

| Dataset | Training | Validation |
|---|---|---|
| **CIFAR10** | 5,000 per class | 1,000 per class |
| **CIFAR100** | 500 per class | 100 per class |
| **ISIC** | 0: Melanoma: 374<br>1: Seborrheic Keratosis: 254<br>2: Others*: 1372 | 0: Melanoma: 30<br>1: Seborrheic Keratosis: 42<br>2: Others*: 78 |
| **IDRiD** | Grad 0: 177<br>Grad 1: 41<br>Grad 2: 195 | Grad 0: 45<br>Grad 1: 10<br>Grad 2: 48 |

\* "Others" means none of the mentioned classes.

To rank the filters, we used a batch of data, which was chosen randomly. The batch size is 256 for CIFAR10 and CIFAR100 and 16 for ISIC and IDRiD datasets. The smaller batch size is used because of varying in models RAM usage as the size of images (256,256,3) for medical datasets is larger than CIFAR datasets (32,32,3).

E. *Pruned Models' Performance*

Firstly, we compared the results of our model for ResNet56 [2] and VGG16 [41] with the best results of state-of-the-art papers including RFP [24], L1 [22], Hrank [11], FilterSketch [23], NISP [25], GAL [26], ABCpruner [27], GCP [28], CLR [29], CHIP [30], SSS [31], Zhao et al. [32], and SWP [18] for the same baseline trained model in Figure 2 to compare the effect of different FLOP reductions in pruning on the accuracy of CIFAR10.

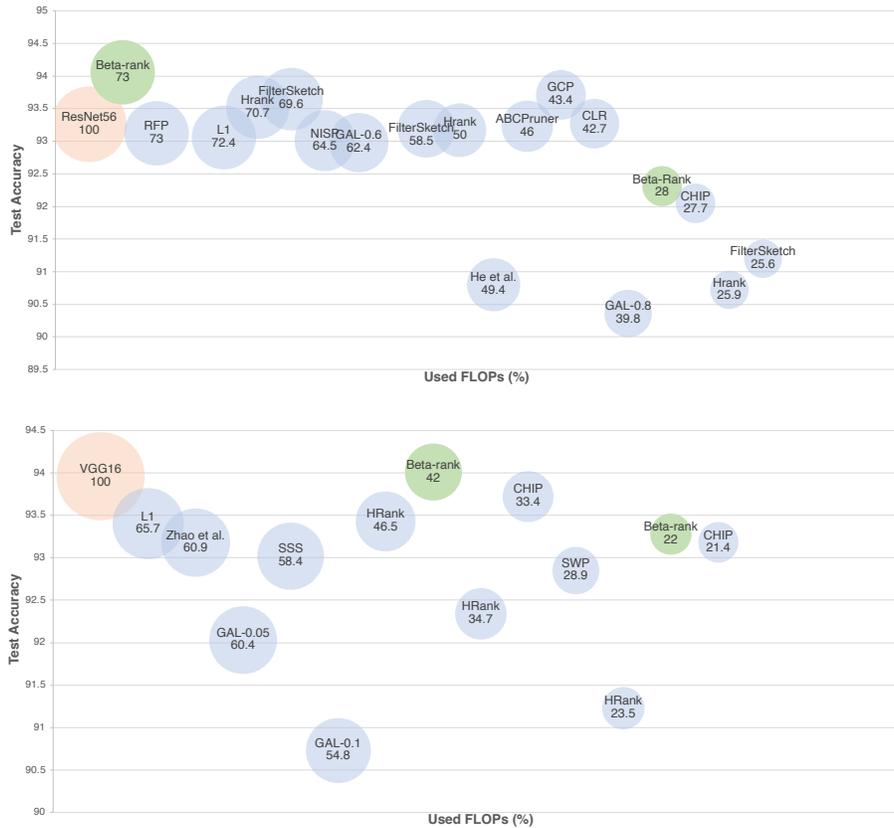

Fig. 2. Bubble plot (top for ResNet56 and bottom for VGG16) illustrating the comparison of various methods' FLOP rates and their impact on accuracy. Each circle represents a pruning method, with the baseline shown in orange. The circle's area indicates the percentage of FLOPs used after pruning compared to the baseline model, while the vertical axis displays the method's accuracy on the CIFAR10 test set.

The results of Figure 2 show that our model has slightly better performance with a higher pruning rate in balanced and standard datasets. While our main contribution is for imbalanced datasets with limited data, the beta fraction has an improvement on balanced datasets as well.

In the next sections, we explore the ranked filters that are reported in Tables II and III. Furthermore, the ranking effect on the understood features will be exposed using visualization.

To ensure the validity of our findings, we conducted each experiment three times while maintaining a constant pruning rate and other training parameters (e.g. number of epochs, and batch size). We tested three methods, namely L1-Norm, Hrank, and Beta-Rank, using three well-known models with different levels of parameters and depth: VGG16 [41], ResNet56, and ResNet110 [2]. As shown in Table 2, the VGG model has a higher parameter-to-FLOPs ratio than the ResNet models. Besides, we included ResNet110 to confirm that our method works for deeper CNNs. However, due to the high variety of models available, it is impossible to cover all of them.

However, we used a constant training procedure (except for the random seeds of some parameters like batch normalizations) to get a mean and standard deviation for each of the pruning rates. The results are provided for three models, including L1-Norm [22] as the backbone of our method, Hrank [11] as one of the state-of-the-art models, and our model. The only difference among the different experiments is their filter rankings and therefore allows for a direct comparison. In addition, the used models as the baseline have been trained from scratch, except for the models used for CIIFAR-10 where the trained weight was available. We trained the baseline models that were not available publicly to the highest possible performance to be very close to the top models. To choose the pruning rate, we used the HRank paper for similar experiments, and the rest rates were selected randomly similar to the same tests. The following tables illustrate the average of the results for different metrics.

*TABLE 2: The mean and std accuracy of methods by repeating each experiment 3 times for CIFAR10 and 100 datasets. FLOPs and parameters are reported for each experiment and rated on a million scale.*

| Model | Dataset | Experiment | Pruning Method | Params Baseline | FLOP Baseline | Accuracy Baseline | Params Punned (↓%) | FLOP Punned (↓%) | Accuracy Punned |
|---|---|---|---|---|---|---|---|---|---|
| VGG-16 | CIFAR-10 | 1 | L1 | 14.98 | 313.73 | 93.96 | 2.76 (81%) | 131.85 (58%) | 93.79 ± 0.15 |
| | | | Hrank | | | | | | 93.68 ± 0.11 |
| | | | Beta-Rank | | | | | | **93.97 ± 0.06** |
| | | 2 | L1 | | | | 1.90 (87%) | 67.50 (78%) | 93.01 ± 0.20 |
| | | | Hrank | | | | | | **93.20 ± 0.09** |
| | | | Beta-Rank | | | | | | 93.13 ± 0.13 |
| | CIFAR-100 | 1 | L1 | 15.04 | 315.21 | 74.24 | 10.48 (30%) | 236.43 (25%) | 73.82 ± 0.16 |
| | | | Hrank | | | | | | **74.02 ± 0.32** |
| | | | Beta-Rank | | | | | | 74.01 ± 0.15 |
| | | 2 | L1 | | | | 6.47 (57%) | 141.49 (55%) | **73.23 ± 0.07** |
| | | | Hrank | | | | | | 72.58 ± 0.32 |
| | | | Beta-Rank | | | | | | 72.75 ± 0.44 |
| ResNet56 | CIFAR-10 | 1 | L1 | 0.85 | 125.49 | 93.26 | 0.66 (22%) | 91.24 (27%) | 93.97 ± 0.14 |
| | | | Hrank | | | | | | 93.70 ± 0.23 |
| | | | Beta-Rank | | | | | | **94.00 ± 0.07** |
| | | 2 | L1 | | | | 0.24 (71%) | 35.37 (72%) | 91.99 ± 0.17 |
| | | | Hrank | | | | | | 91.91 ± 0.25 |
| | | | Beta-Rank | | | | | | **92.09 ± 0.19** |
| | CIFAR-100 | 1 | L1 | 0.86 | 127.63 | 73.120 | 0.75 (13%) | 107.94 (15%) | **73.09 ± 0.09** |
| | | | Hrank | | | | | | 72.62 ± 0.15 |
| | | | Beta-Rank | | | | | | 72.93 ± 0.29 |
| | | 2 | L1 | | | | 0.58 (67%) | 84.87 (33%) | 66.77 ± 0.12 |
| | | | Hrank | | | | | | 66.67 ± 0.26 |
| | | | Beta-Rank | | | | | | **67.11 ± 0.12** |

The results of Table 2 show that the $\beta$-rank model slightly improved the results of the L1-Norm method, in most of the experiments. However, the advantage of the presented model can be shown on more challenging datasets.

For the medical datasets, we used ResNet56 and ResNet110 and avoided training models that have a high number of parameters. Table 3 reports the results of these experiments.

TABLE 3: The mean (grey rows) and std (below the grey rows) results of methods by repeating each experiment 3 times for ISIC and IDRiD datasets. FLOPs and parameters are rated on a million scale.

| Dataset & Model | ISIC & ResNet56 | | | | | |
|---|---|---|---|---|---|---|
| FLOP (↓%) | 8167.62 | 6242.63 (24%) | | | 4781.00 (41%) | |
| Params (↓%) | 0.85 | 0.66 (22%) | | | 0.55 (35%) | |
| Pruning method | Baseline | L1 | Hrank | Beta-Rank | L1 | Hrank | Beta-Rank |
| Accuracy | 73.33 | 69.56 | 59.56 | **73.56** | 60.89 | 57.33 | **72.44** |
| | | 4.91 | 3.01 | 1.68 | 3.67 | 2.40 | 1.68 |
| Precision | 66.5 | 62.07 | 44.57 | **69.53** | 58.80 | 45.37 | **66.30** |
| | | 1.55 | 7.70 | 2.32 | 8.45 | 10.35 | 3.83 |
| Recall | 63.2 | 56.60 | 46.97 | **64.70** | 49.93 | 43.30 | **62.37** |
| | | 4.16 | 3.31 | 1.35 | 5.73 | 3.18 | 3.28 |
| Specificity | 85.5 | 81.13 | 72.93 | **84.00** | 75.10 | 72.20 | **83.17** |
| | | 3.59 | 1.43 | 0.98 | 2.96 | 1.54 | 1.40 |
| Dataset & Model | ISIC & ResNet110 | | | | | |
| FLOP (↓%) | 16453.47 | 9093.39 (45%) | | | 12064.74 (27%) | |
| Params (↓%) | 1.73 | 1.05 (39%) | | | 1.33 (23%) | |
| Pruning method | Baseline | L1 | Hrank | Beta-Rank | L1 | Hrank | Beta-Rank |
| Accuracy | 72.66 | 66.67 | 67.78 | **73.33** | 62.00 | 59.11 | **71.56** |
| | | 0.67 | 2.69 | 0.67 | 2.67 | 1.39 | 0.39 |
| Precision | 69.7 | 57.57 | 56.10 | **63.27** | 46.57 | 44.10 | **63.43** |
| | | 3.75 | 4.00 | 3.44 | 4.98 | 4.26 | 2.46 |
| Recall | 65.7 | 57.90 | 57.43 | **64.40** | 50.83 | 46.87 | **63.93** |
| | | 2.27 | 2.89 | 3.77 | 4.45 | 2.86 | 0.81 |
| Specificity | 82.6 | 78.47 | 79.00 | **84.60** | 75.03 | 74.17 | **81.67** |
| | | 1.24 | 2.54 | 1.39 | 2.31 | 2.12 | 0.91 |
| Dataset & Model | IDRiD & ResNet56 | | | | | |
| FLOP (↓%) | 8167.62 | 5243.91 (36%) | | | 4422.36 (46%) | |
| Params (↓%) | 0.85 | 0.57 (33%) | | | 0.49 (42%) | |
| Pruning method | Baseline | L1 | Hrank | Beta-Rank | L1 | Hrank | Beta-Rank |
| Accuracy | 82.52 | 73.79 | 71.52 | **79.94** | 75.08 | 70.87 | **80.58** |
| | | 0.97 | 1.48 | 1.12 | 1.12 | 0.97 | 1.68 |
| Precision | 55.4 | 54.33 | 53.20 | **65.40** | 57.53 | 54.80 | **61.37** |
| | | 3.01 | 2.02 | 1.15 | 2.90 | 2.75 | 7.51 |
| Recall | 61.3 | 59.47 | 56.73 | **69.67** | 62.60 | 58.10 | **65.60** |
| | | 3.33 | 1.07 | 1.50 | 2.71 | 2.69 | 6.12 |
| Specificity | 89.5 | 83.97 | 82.67 | **88.07** | 84.80 | 82.10 | **88.27** |
| | | 0.81 | 0.93 | 0.40 | 0.72 | 0.95 | 1.14 |
| Dataset & Model | IDRiD & ResNet110 | | | | | |
| FLOP (↓%) | 16453.47 | 9093.39 (45%) | | | 6839.93 (58%) | |
| Params (↓%) | 1.73 | 1.05 (39%) | | | 0.80 (56%) | |
| Pruning method | Baseline | L1 | Hrank | Beta-Rank | L1 | Hrank | Beta-Rank |
| Accuracy | 79.61 | 73.14 | 78.32 | **80.26** | 74.76 | 61.49 | **81.88** |
| | | 1.48 | 1.12 | 1.12 | 2.57 | 2.44 | 1.48 |
| Precision | 0.652 | 53.27 | 62.27 | **64.27** | 62.07 | 46.67 | **73.60** |
| | | 2.84 | 4.005 | 1.39 | 7.05 | 2.35 | 14.08 |
| Recall | 0.676 | 58.33 | 65.47 | **66.60** | 64.93 | 47.70 | **68.63** |
| | | 1.51 | 1.79 | 1.18 | 6.05 | 3.12 | 4.52 |
| Specificity | 0.876 | 84.63 | 86.97 | **88.67** | 85.20 | 76.50 | **89.20** |
| | | 0.81 | 0.93 | 0.40 | 0.72 | 0.95 | 1.14 |

Results in Table 3 demonstrated that in a constant training environment, pruning using the Beta-Rank method outperformed other methods by a substantial margin.

### F. Pruning Impact on Mobile Device Resources

We developed an Android application to test the effect of parameter reduction and FLOPs on execution time and memory consumption. We used the Android version 12.0 and our tested device was a Samsung Galaxy

A31 with the same Android version, 4GB of memory, and 1.7 GHz of processor frequency clock. Table 4 shows the difference between baseline models and pruned models based on $\beta$-rank in real-world settings for repeating each prediction five times.

TABLE 4: *Results of real-world experiments using the developed Android app for measuring time (in milliseconds) and Memory (in Megabytes).*

| Model (FLOPs pruning rate) | Dataset | Baseline (Mean ± Std) | | Pruned (Mean ± Std (↓%)) | |
|---|---|---|---|---|---|
| | | Time | Memory | Time | Memory |
| ResNet-56 (27%) | CIFAR-10 | 411.8 ± 13.29 | 0.216 ± 0.005 | **400.2 ± 9.12 (2.8)** | **0.190 ± 0.016 (12.0)** |
| Vgg-16(78%) | CIFAR -10 | 617.2 ± 6.870 | 0.778 ± 0.321 | **166.8 ± 3.033 (72.9)** | **0.158 ± 0.016 (79.6)** |
| ResNet-56 (15%) | CIFAR -100 | 468.2 ± 27.79 | 0.276 ± 0.188 | **438.6 ± 24.183 (6.3)** | **0.226 ± 0.083 (18.1)** |
| Vgg-16 (25%) | CIFAR -100 | 616.0 ± 2.739 | 0.994 ± 0.084 | **450.8 ± 9.230 (26.8)** | **0.460 ± 0.288 (53.7)** |
| ResNet-56 (36%) | IDRiD | 1322.4 ± 12.012 | 0.314 ± 0.206 | **1306.8 ± 261.99 (1.1)** | **0.200 ± 0.030 (36.3)** |
| ResNet-110 (58%) | IDRiD | 2244.0 ± 12.12 | 0.440 ± 0.051 | **1789.0 ± 22.82 (20.2)** | **0.382 ± 0.034 (13.1)** |
| ResNet-56 (41%) | ISIC | 969.4 ± 27.42 | 0.160 ± 0.031 | **794.6 ± 5.68 (18.0)** | **0.142 ± 0.044 (11.2)** |
| ResNet-110 (27%) | ISIC | 2231.6 ± 11.58 | 0.410 ± 0.101 | **2022.2 ± 14.92 (9.4)** | **0.338 ± 0.115 (17.5)** |

As can be seen from Table 4 the pruned models can save up to 72% of the time and up to 53% of memory compared with baseline models.

*G. Ranking Stability Analysis*

One of the main concerns of filter pruning methods is the reproducibility of filter ranking. In other words, a pruning method may select specific filters as top rank and other filters as low rank based on a random batch of input samples. If the method is robust and reproducible, it is expected that the same filters have been selected based on a new random batch of input samples. To further explore this issue, we chose the top and least 25% of each layer's ranked filters and explored the number of non-repetitive choices of each presented method in ResNet56, which has been trained on all datasets separately. For example, for a layer with 16 filters, the top 25% is 4 filters. If the model in three repetitions of ranking based on different samples chooses different filters in each experiment, it means 12 different filters will be selected, which is the worst scenario of filter selection. In the best scenario, the method will select the same four filters in all three ranking repetitions based on different samples. The fraction of the selected number of filters to the worst possible number will give us a criterion for evaluating which methods give the most stable results. Figure 3 shows the top and least 25% ranked filters for all layers of the ResNet56 model.

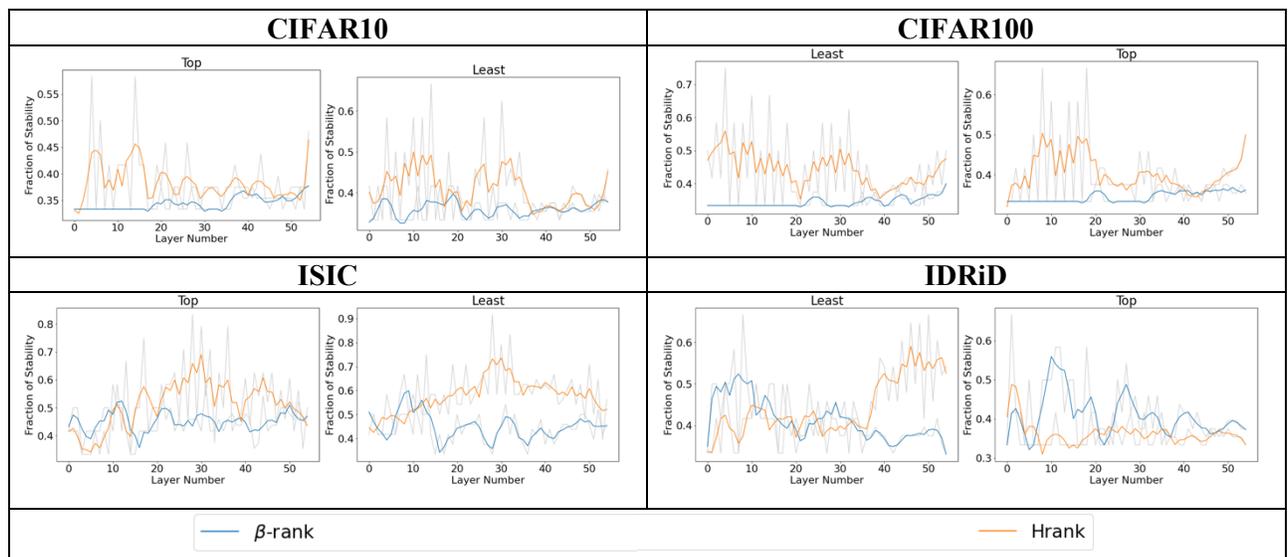

Fig. 3. *Stability of filter selection using different filter pruning methods for ResNet56. The actual values are presented in grey color and smoothed versions of them are presented in blue and orange. X-axes show the layers number of the model which is 56 and Y-axes show the fraction of stability for the top and least 25% of filters. The best values will be smaller values for each layer.*

The visualization of two models for the two rich and balanced datasets of CIFAR10 and CIFAR100 demonstrates that both models entail a stable behavior in filter selection whereas, in the challenging medical datasets, the values experience higher variation and more erratic behavior. Moreover, $\beta$-rank shows a downward curve with increasing layers for the two last datasets illustrating a tendency for more stable selection. This behavior is natural as the first layers of CNNs extract lower-level features and as the layers go deeper, features become higher in the level of information (e.g. mouth and eyes are high-level features compare with the edge in a face image). Therefore, for input images that are not detailed, CNN models can find the edge using different filters each time and this can be the main reason for instability at the first layers.

*H. Heatmap Visualization*

One of our objectives is to explore the effect of filter pruning on choosing the most relevant features from the input image. In general, convolutional models find the appropriate features in the input image that are related to the final target. We want to know, does filter pruning reduce the number of features in the input image, or it is better to reduce high-level information from the last layers of networks when we prune a model?

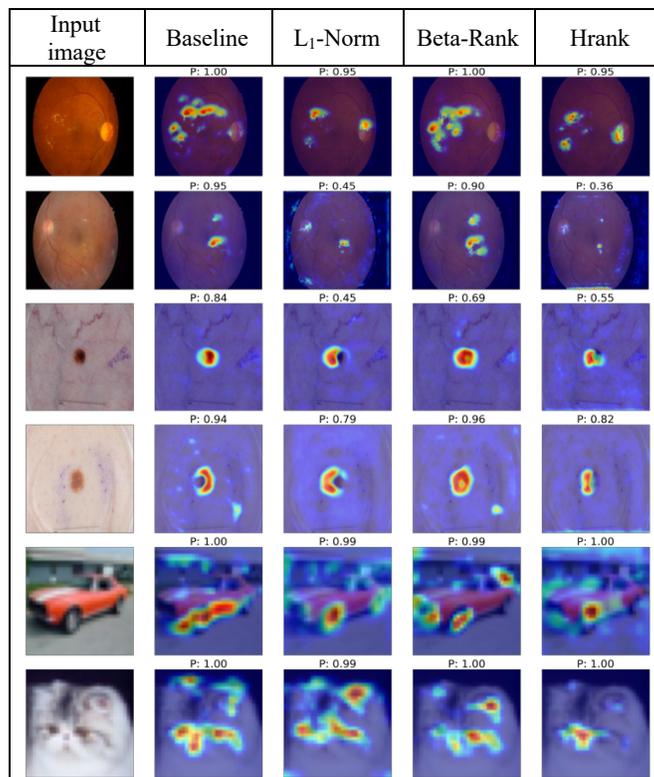

*Fig. 4. GradCam visualization for different datasets. The shown value above each image stands for the probability of the correct class prediction. P stands for the probability of prediction for the correct class.*

Figure 4 shows the visualization of the GradCam [42] model for a constant pruning rate for all pruned models. Those networks that classified more classes or used the full capacity of the network tried to limit the features to get the same results. However, for the medical datasets, the pruned models have more capacity to prune and the feature are intact or even extended. However, as presented different methods focus on different features to predict their results. For example, in the cat image of Figure 4, the baseline model considered eyes, ears, nose, and mouth for its prediction. But $\beta$-rank model had less focus on one of the eyes and less emphasis on the ears to predict the result. By considering this point the pruning rates for all methods are the same, it can be seen $\beta$-rank can cover more efficient features that might be more general in images that are more complex to predict.

## DISCUSSION

In this work, we present a novel pruning method that can have equal performance (Table 2) with state-of-the-art methods on balanced and rich datasets and tested on potential real-world applications demonstrating reduced execution time and memory utilization. Reducing memory usage in deploying a machine-learning model is advantageous because it not only reduces the cost of operating the model but also improves the speed at which the model can be deployed, as less memory equates to faster operation. Moreover, when a reduced amount of power is required, it results in a more efficient model. Having these capabilities could extend the use of larger models that can be deployed on constrained hardware such as edge devices.

As can be seen in Table 2, by comparing the accuracy results of the CIFAR10 dataset and ResNet56 for the three methods, the reported mean values were not significantly different since the standard deviations of the three methods overlap. Therefore, it is not possible to prioritize one method over the other.

Despite these difficulties, our method can cope with imbalanced datasets (Table 3) that have a limited number of samples by retaining the useful information that state-of-the-art methods failed to operate accurately when target datasets had limited numbers of samples. According to the results presented in Table 3, the $\beta$-rank method produces significant differences in results compared to other methods, based on their reported standard deviations. Due to the small size of the datasets (Table 1), methods presented higher standard deviations for the small datasets than standard datasets (e.g. CIFAR10). Consequently, the presented methods in Figure. 3 produce more variability in their results when a small sample size is used.

To ensure a fair comparison of all models, we conducted experiments with constant hyperparameters and pruning rates (Tables II and III) and repeated them three times. We then used the mean and standard deviation to demonstrate the differences in performance between the models, rather than just focusing on the most relevant results.

To demonstrate the real-world capability of our method, we have developed an Android application that assesses the real-world performance of pruned models with baselines using execution time and memory utilization (Table 4). As it appears in the real world, reducing the number of FLOPs may result in further reductions in execution time and memory utilization. This is because the device could have other applications that result in more or fewer reductions in memory utilization.

As part of our analysis, we compared the ranking stability of our method with Hrank, which ranks filters based on a batch of data. As can be seen in Figure 3, the stability fractions have a smaller number that results in more stable behavior and is less dependent on random samples than the Hrank method.

The presented GradCam visualization (Figure 4) illustrates how filter pruning methods can consider useful information and omit unnecessary ones from the input image to produce the same results with fewer parameters and computational costs. The car image in Figure 4 demonstrates this claim clearly since pruning methods focus on the main information for car classification, such as wheels, windows, and lights. These features were more accurate than the baseline highlighted features.

Despite the listed advantages, there are also some limitations. The pruning rate for each layer of baseline networks is manually set for this work, as well as previous works. This assignment may not be the optimal one, and some layers may work with fewer filters while others may not. Furthermore, we employed three different models and four datasets in our evaluation. However, a general approach to filter pruning may require assessments of a wide variety of models and datasets to ensure generalizability.